\pdfoutput=1

\documentclass[11pt]{article}
\usepackage{preamble}

\title{Solving the Challenge Set without Solving the Task: \\ On Winograd Schemas as a Test of Pronominal Coreference Resolution}

\author{Ian Porada \\
  Mila - Quebec AI Institute \\
  McGill University \\
  {\tt ian.porada@mail.mcgill.ca}
  \And
  Jackie Chi Kit Cheung \\
  Mila - Quebec AI Institute \\
  McGill University \\
  Canada CIFAR AI Chair \\
  {\tt jackie.cheung@mcgill.ca}
}

\begin{document}
\maketitle
\begin{abstract}
Challenge sets such as the Winograd Schema Challenge (WSC) are used to benchmark systems' ability to resolve ambiguities in natural language. If one assumes as in existing work that solving a given challenge set is at least as difficult as solving some more general task, then high performance on the challenge set should indicate high performance on the general task overall. However, we show empirically that this assumption of difficulty does not always hold. In particular, we demonstrate that despite the strong performance of prompted language models (LMs) on the WSC and its variants, these same modeling techniques perform relatively poorly at resolving certain pronominal ambiguities attested in OntoNotes and related datasets that are perceived to be easier. Motivated by these findings, we propose a method for ensembling a prompted LM with a supervised, task-specific system that is overall more accurate at resolving pronominal coreference across datasets. Finally, we emphasize that datasets involving the same linguistic phenomenon draw on distinct, but overlapping, capabilities, and evaluating on any one dataset alone does not provide a complete picture of a system's overall capability.
\end{abstract}

\section{Introduction}
\label{sec:introduction}

The Winograd Schema Challenge~\citep[WSC;][]{levesque2012winograd} is a challenge set of ambiguous pronominal coreference resolution (PCR) problems, one of many popular challenge sets used to evaluate NLP systems \citep[\textit{e.g.,}][]{isabelle-etal-2017-challenge,clark2018think,mccoy-etal-2019-right}. Challenge sets are constructed to consist of relatively difficult instances of some more general task. In many cases, systems' performance on challenge sets is considered in isolation of performance on the broad range of ambiguous expressions attested in natural corpora on which the general task being studied could also be evaluated;\footnote{We use the term \textit{natural corpora} to refer to text that was not explicitly constructed or elicited for research purposes. An \textit{attested} expression is one appearing in natural corpora in contrast to \textit{constructed} expressions that commonly compose challenge sets.} \textit{e.g.,} systems are often evaluated on the WSC without considering how those same systems might perform on a diverse range of attested pronominal expressions~\citep[][\textit{i.a.}]{kocijan-etal-2019-surprisingly,shen-etal-2021-unsupervised,eval-harness,achiam2023gpt}.

\begin{figure}[t] 
\centering
  \includegraphics[width=0.48125\textwidth]{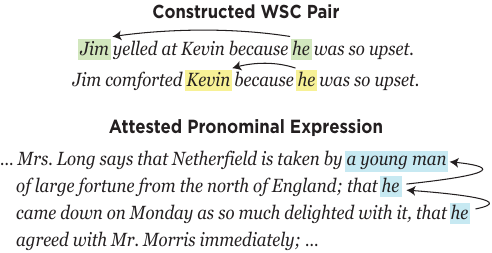}
  \caption{
  \textit{Top}: An example minimal pair from the WSC. \textit{Bottom}: Pronouns attested in the novel \textit{Pride and Prejudice} and annotated for coreference by \citet{vala-etal-2016-annotating}.
  }
  \label{fig:intro-example}
  \vspace{-5pt}
\end{figure}

The WSC specifically consists of minimal pairs of sentences, each containing an ambiguous pronoun (Figure~\ref{fig:intro-example}). These pairs are manually constructed such that consistently disambiguating the pronouns is believed to require the types of commonsense world knowledge and reasoning ability a human reader might rely on. Considering the recent success of language model (LM) based approaches at resolving WSC instances, some of the original authors of the WSC have declared the challenge set solved~\citep{Kocijan_2023}.

And yet, in this work we demonstrate that the same LM-based systems that have reportedly solved the WSC and its variants are relatively inaccurate at resolving certain ambiguous pronominal expressions attested in natural corpora and annotated in OntoNotes~\citep{hovy2006ontonotes} and related datasets. One may find this result surprising given that ``the point of the WSC is to test programs that claim to have solved the problem of pronoun reference resolution''~\citep{Kocijan_2023} and that WSC instances are believed to represent relatively difficult examples of PCR~\citep{peng-etal-2015-solving}.

We specifically consider prompted LMs as systems that are relatively accurate at resolving Winograd schemas; among LMs, our experiments focus on the Llama family of models~\citep{touvron2023llama,dubey2024llama3herdmodels}, although we present evidence that our results generalize across LM families including OLMo~\citep{groeneveld-etal-2024-olmo} and Mistral~\citep{jiang2023mistral}. We compare the performance of prompted LMs up to 70B parameters against state-of-the-art coreference resolution systems, such as Maverick~\citep{martinelli-etal-2024-maverick}, which are known to be accurate at resolving attested pronominal coreferences.

We evaluate systems across 11 datasets. Six of these datasets contain PCR problems for text attested in natural corpora, \textit{e.g.,} OntoNotes 5.0~\citep{ontonotes5} and OntoGUM~\citep{zhu-etal-2021-ontogum}. The other five datasets consist of PCR problems that were constructed for WSC-like challenge sets, \textit{e.g.,} Winogrande~\citep{sakaguchi2021winogrande} and DPR~\citep{rahman-ng-2012-resolving}.

When comparing against unsupervised baselines, we find LMs are generally more accurate across all datasets; however, \textit{whereas supervised coreference resolution models perform relatively poorly on the WSC, we find these same systems are more accurate than prompted LMs at resolving certain attested pronouns}. This finding is consistent across test sets of diverse annotation guidelines and textual genres. 

Motivated by these results, we propose a method for ensembling a prompted LM with a task-specific system in order to achieve a final system that is overall more accurate at resolving pronominal coreference across datasets. This ensembling method functions by heuristically determining salient discourse entities for which coreference is disambiguated by a state-of-the-art coreference resolution system trained on OntoNotes. Meanwhile, the remaining instances are disambiguated using an LM prompted with in-context examples. In most cases, the final system is more accurate at resolving pronouns occurring in attested expressions and WSC-like challenge sets.

Ultimately, our findings illustrate the point that datasets involving the same linguistic phenomenon draw on distinct, but overlapping, capabilities; therefore, no one dataset alone is capable of providing a complete picture of a system's overall performance. We therefore argue that challenge set results should be considered in conjunction with results on evaluations that encompass a diverse range of attested expressions.

\paragraph{Contributions.}

Our primary contributions can be summarized as follows:
\vspace{-3pt}
\begin{enumerate}
\itemsep0em 
    \item We formalize and empirically question \textit{the challenge set assumption} that solutions to a challenge set generalize to diverse, attested instances of the phenomenon being targeted. In the case of PCR, we provide direct evidence that this assumption does not hold.
    \item We present a formatted collection of 11 datasets that follow the same, consistent formulation of PCR. Using this collection, we evaluate and compare multiple types of approaches to PCR including supervised models, prompted LLMs, and rule-based systems.
\end{enumerate}

\section{Related Work}
\label{sec:related-work}

PCR is broadly the task of determining which linguistic expressions refer to the same discourse entity as a given pronominal expression \citep{hobbs1978resolving}. See \citet{zhang-etal-2021-brief} and \citet{poesio2023computational} for related surveys.

Proposed systems for resolving pronominal coreference have traditionally relied on heuristic rules often in combination with unsupervised statistical patterns of handcrafted features~\citep[][\textit{i.a.}]{poon-domingos-2008-joint,charniak-elsner-2009-em,raghunathan-etal-2010-multi,lee-etal-2011-stanfords}. More recently, LM-based approaches have been proposed including: LMs finetuned on supervised training data~\citep{zhang-etal-2019-knowledge,zhao-etal-2022-pcr4all}, weakly supervised LMs~\citep{kocijan-etal-2019-wikicrem,shen-etal-2021-unsupervised}, and prompting LMs by formatting PCR as either a cloze task~\citep{trinh2018simple,radford2019language} or question answering~\citep{brown2020language,wang-etal-2022-super,le2023large,zhu-etal-2024-large}.

The ability of a system to perform PCR has been evaluated generally on: 1) collections of ambiguous pronouns attested in natural text~\citep{hobbs1978resolving,lappin-leass-1994-algorithm,webster-etal-2018-mind}, 2) the subsets of larger coreference resolution datasets that include pronominal coreference~\citep{martschat-strube-2014-recall,zhang-etal-2019-knowledge,lu-ng-2020-conundrums}, and 3) challenge sets composed of WSC-like instances~\citep{rahman-ng-2012-resolving,emami-etal-2019-knowref,sakaguchi2021winogrande}.

The WSC and inspired datasets have been adopted by researchers studying the more general task of coreference resolution to be used as challenge sets in addition to more canonical evaluations such as OntoNotes~\citep{peng-etal-2015-solving,toshniwal-etal-2021-generalization,zhao-etal-2022-pcr4all}. Such work has shown that systems designed for coreference resolution perform poorly on the WSC. We adopt the perspective of this line of work and view WSC-like datasets as challenge sets of PCR.

Recent advances in language modeling have proven accurate at resolving WSC instances when compared to earlier approaches, in some cases nearing approximates of human accuracy~\citep{brown2020language,wei2022finetuned,touvron2023llama}. However, similar techniques have been shown to be less accurate than supervised models when evaluated on established evaluations of the general task of coreference resolution~\citep{yang-etal-2022-gpt,le2023large,zhu-etal-2024-large,gan-etal-2024-assessing-capabilities}. Our work diverges from these studies by focusing specifically on PCR rather than the broader concept of coreference which has multiple competing definitions~\citep{recasens-hovy-2010-coreference,can_we_fix}.

\section{Method}
\label{sec:method}

In this section, we formulate the problem of pronominal coreference resolution (PCR) and provide a high-level description of how system accuracy is evaluated. We also formalize the assumptions commonly made when evaluating on challenge sets so that we can explicitly test if these assumptions hold.

\subsection{Problem Formulation}

We consider the task of PCR formulated as follows: given a text passage $w = (w_1, \ldots, w_t)$, resolve some pronominal expression $x$ to its correct antecedent $a$, where $x$ and $a$ are subspans of $w$.
We study a restricted version of this problem formulated as binary classification.

\begin{figure}[t]
\centering
  \includegraphics[width=0.48125\textwidth]{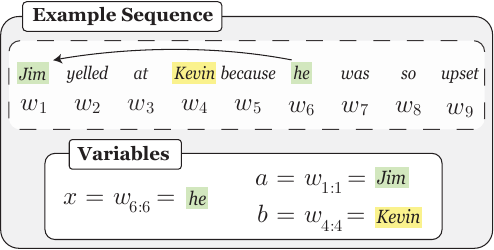}
  \caption{
  An example instance and the corresponding variables: the pronoun $x$, antecedent $a$, and distractor candidate $b$.
  }
  \label{fig:formulation-example}
\end{figure}

More explicitly, we assume that exactly one of two candidate antecedents in $w$ is the correct resolution of $x$. This formulation accommodates both WSC-style and datasets containing annotations of coreference in occurring in natural corpora. Formally, given $w$, $x$, and a set of two candidate antecedents $\{\hat{a}_1, \hat{a}_2 \}$, the goal is to correctly determine which candidate antecedent corresponds to the true antecedent $a = w_{k:l}$. The other candidate is some distractor mention $b = w_{m:n}$ that refers to a discourse entity but does not corefer with $x$. An example instance is given in Figure~\ref{fig:formulation-example}.

\subsection{Challenge Set Assumptions}

An assumption of the WSC is that solving WSC instances is more difficult than resolving other instances of the PCR task such as pronouns attested in natural corpora. We formulate this assumption as follows (Def.~\ref{def:the-challenge-set-assumption}). We will then test this assumption empirically by comparing the performance of various systems across both challenge set instances and attested pronouns.

To premise, let $C$ be a challenge set and $D$ some other dataset representing the same task. Furthermore, let $\theta$ and $\phi$ be systems that are to be evaluated based on their performance on the given task. Function $U$ represents a measure of the performance of a system on a given dataset, \textit{e.g.,} the performance of $\theta$ on $C$ is measured as $U(\theta, C)$.

\begin{definition}[The Challenge Set Assumption]
\label{def:the-challenge-set-assumption}
The ordering of model performance on the challenge set $C$ is preserved on dataset $D$. That is, $U(\theta, C) > U(\phi, C) \implies U(\theta, D) > U(\phi, D)$.
\end{definition}

Intuitively, the assumption is that because $C$ is strictly more difficult than $D$, systems that are relatively accurate on $C$ should be relatively accurate on $D$ as well.

\begin{figure*}[t]
\centering
  \includegraphics[width=0.95\textwidth]{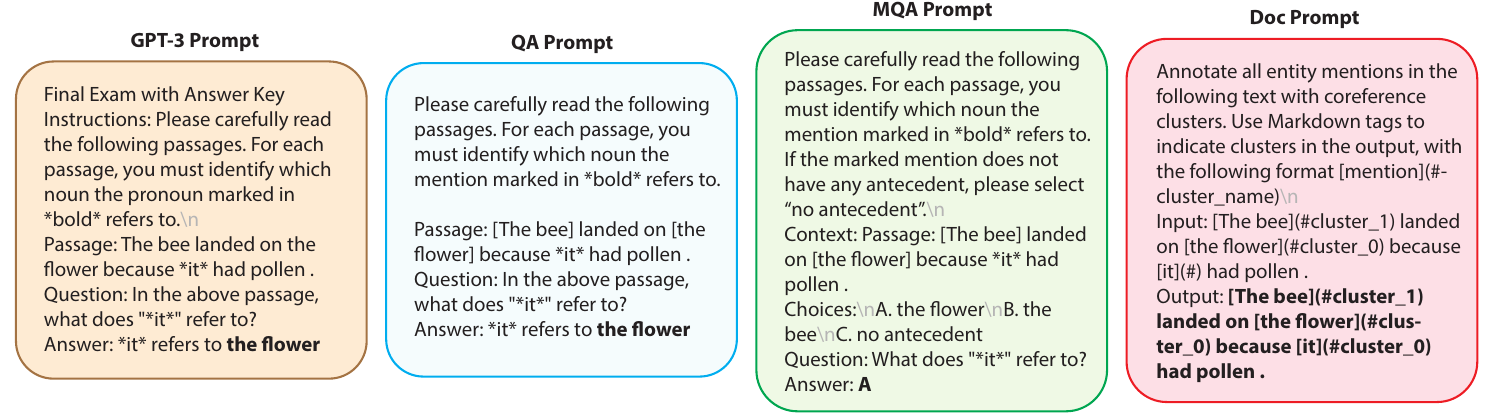}
  \caption{A training set instance from the Definite Pronoun Resolution (DPR) dataset~\citep{rahman-ng-2012-resolving} formatted using each of the corresponding prompts. Denoted in bold is the expected model output. The GPT-3 prompt~\citep{brown2020language} does not rely on gold mention span annotations. QA Prompt and Doc Prompt were presented by \citet{le2023large}. The multiple-choice QA (MQA) prompt was presented by \citet{zhu-etal-2024-large}.
  }
  \label{fig:prompts}
\end{figure*}

\subsection{Evaluating Performance}

To test this assumption, we evaluate systems across multiple test sets. Here we describe how the performance function $U$ is calculated.

\paragraph{Attested Pronominal Expressions.} To evaluate performance on attested pronominal expressions, we start with some existing dataset of identity coreference relations annotated in datasets of curated natural corpora. We then take all mentions that are in a coreference relation and are within a predefined set of pronouns to be a coreferring pronominal expressions $x$.\footnote{The following strings are considered as pronouns: "she", "her", "he", "him", "them", "they", "it", "his", "hers", "its", "their", "theirs", "this", "that", "these", "those".} We take the text passage $w$ to be the concatenation of the sentence in which $x$ occurs with the preceding two sentences. For each $x$ for which a single coreferring nominal antecedent mention $a$ occurs in the context $w$, and for which at least one coreferring expression $b$ that does not corefer with $x$ also occurs in $w$, we create a test instance. In the event that there are multiple candidates that could be chosen as $b$, we randomly sample one. We measure performance based on a system's accuracy at resolving these instances.

This formulation and the predefined set of pronouns follow the conventional setup for PCR used in existing work~\citep{yang-etal-2003-coreference,ng2005supervised,li-etal-2011-pronoun,zhang-etal-2019-knowledge,zhao-etal-2022-pcr4all}. App.~\ref{sec:evaluation-differences} provides further details regarding how datasets are formatted.

\paragraph{WSC-like Challenge Sets.} WSC instances generally follow the formulation of PCR we have outlined above---\textit{i.e.,} the basic premise of a WSC is that there is some text $w$ containing a pronoun $x$ with two candidate antecedents $\{\hat{a}_1, \hat{a}_2 \}$---so we perform minimal formatting of existing WSC-like challenge sets so that examples are in the same form as for attested datasets. This requires tokenization and in certain cases determining the exact span of candidate mentions. Further details are provided in App.~\ref{sec:evaluation-differences}. We can then directly compute accuracy as the ratio of instances where the system predicts the correct candidate antecedent.

\section{Experiments}
\label{sec:experiments}

In this section, we describe our experimental setup in detail which tests whether the challenge set assumption holds empirically. We compare prompted LMs that are known to be accurate at the WSC against task-specific systems known to be accurate at resolving certain attested pronouns (\S\ref{sec:models}). Systems are evaluated across 11 datasets spanning both attested and WSC-like instances (\S\ref{sec:datasets}).

\subsection{Systems}
\label{sec:models}

\subsubsection{Prompted Language Models}

In recent years, prompted LMs have proven accurate at the WSC. One would therefore expect such systems to be relatively accurate at resolving attested pronominal expressions if the challenge set assumption holds. Prompted LMs function by predicting the correct antecedent span $a$ given a problem instance $(w, x, \{\hat{a}_1, \hat{a}_2 \})$ which is formatted using a particular textual prompt template that may possibly include in-context examples.

\paragraph{Llama 3.1} As a prompted LM we focus on the Llama 3.1 family of models at various sizes~\citep{dubey2024llama3herdmodels}. These are competitive open-weights LMs. We consider either the base or instruct version as specified in each experiment. The instruct versions were additionally finetuned on instruction-tuning data, such as the Flan collection \citep{pmlr-v202-longpre23a}, and human preference annotations. We evaluate the 8B and 70B parameter model sizes.

In the experiments where we consider few-shot prompted Llama models, we also compare against a supervised Llama 3.1 8B model which we finetune to resolve WSC by training on public training sets formatted using a QA prompt.

\paragraph{Additional LMs} We additionally compare performance against the smaller Llama 3.2 models, the fully open source OLMo model~\citep{groeneveld-etal-2024-olmo}, and the Mistral-NeMo 12B parameter model~\citep{mistralNemo}.

\paragraph{Prompting Techniques.}

Our goal is not to propose new prompting techniques, so we experiment using four existing prompt templates sourced from the literature. These templates are shown in Figure~\ref{fig:prompts}. The GPT-3 prompt was used by \citet{brown2020language} for evaluating GPT-3 on the SuperGLUE WSC~\citep{wang2019superglue} and does not require gold mention annotations. For this prompt, we check the string match of the model output as in \citet{brown2020language}. The additional prompts (QA, MQA, and Doc prompts) were proposed for using language models to explicitly perform the task of coreference resolution and do require explicit candidate mention spans. For these prompts, of the candidate outputs, we take that with the highest probability assigned by the language model to be the model prediction. Another common approach is to formulate WSC-like instances as a cloze-task~\citep{trinh2018simple,eval-harness}. We do not consider this prompting technique, however, as it is not compatible with pronominal references whose resolution depends on the grammatical features of the pronoun being considered.

We evaluate prompted LMs in zero- and few-shot settings depending on what comparison is being made. In the zero-shot setting, the LM is only prompted with the corresponding instructions and input passage. In the few-shot setting, we use instruction-tuned version of the Llama 3.1 models with 32 training instances prepended to the input.

\subsubsection{Task-Specific Systems}

We compare the performance of prompted LMs against the following task-specific systems designed for the general problem of coreference resolution. Such coreference resolution models are believed to perform poorly on the WSC. One would therefore expect prompted LMs to outperform these task-specific systems across all PCR datasets given \textit{the challenge set assumption} (Def.~\ref{def:the-challenge-set-assumption}).

\paragraph{dcoref} As a representative unsupervised system, we consider the ``Stanford Deterministic Coreference Resolution System''~\citep[dcoref;][]{lee-etal-2013-deterministic}. This is a deterministic, rule-based approach to the general problem of identity coreference resolution and does not rely on supervised examples of coreference relations. The system is optimized to perform well on the OntoNotes dataset. This system uses 10 sieves (such as string match and grammatical feature agreement) to identify potentially coreferring mentions. We use the most recent version implemented in Stanford Core NLP \citep{manning-etal-2014-stanford}. Around 30 percent of OntoNotes errors are described as pronominal anaphora errors in the original dcoref paper.

\paragraph{Maverick} As a representative example of a supervised system, we consider the state-of-the-art Maverick coreference resolution system~\citep{martinelli-etal-2024-maverick}. We use the publicly released weights of the best performing system which consists of a DeBERTa-v3 encoder~\citep{he2021debertav3} finetuned on OntoNotes.

\begin{figure*}[t]
\centering
  \includegraphics[width=0.99\textwidth]{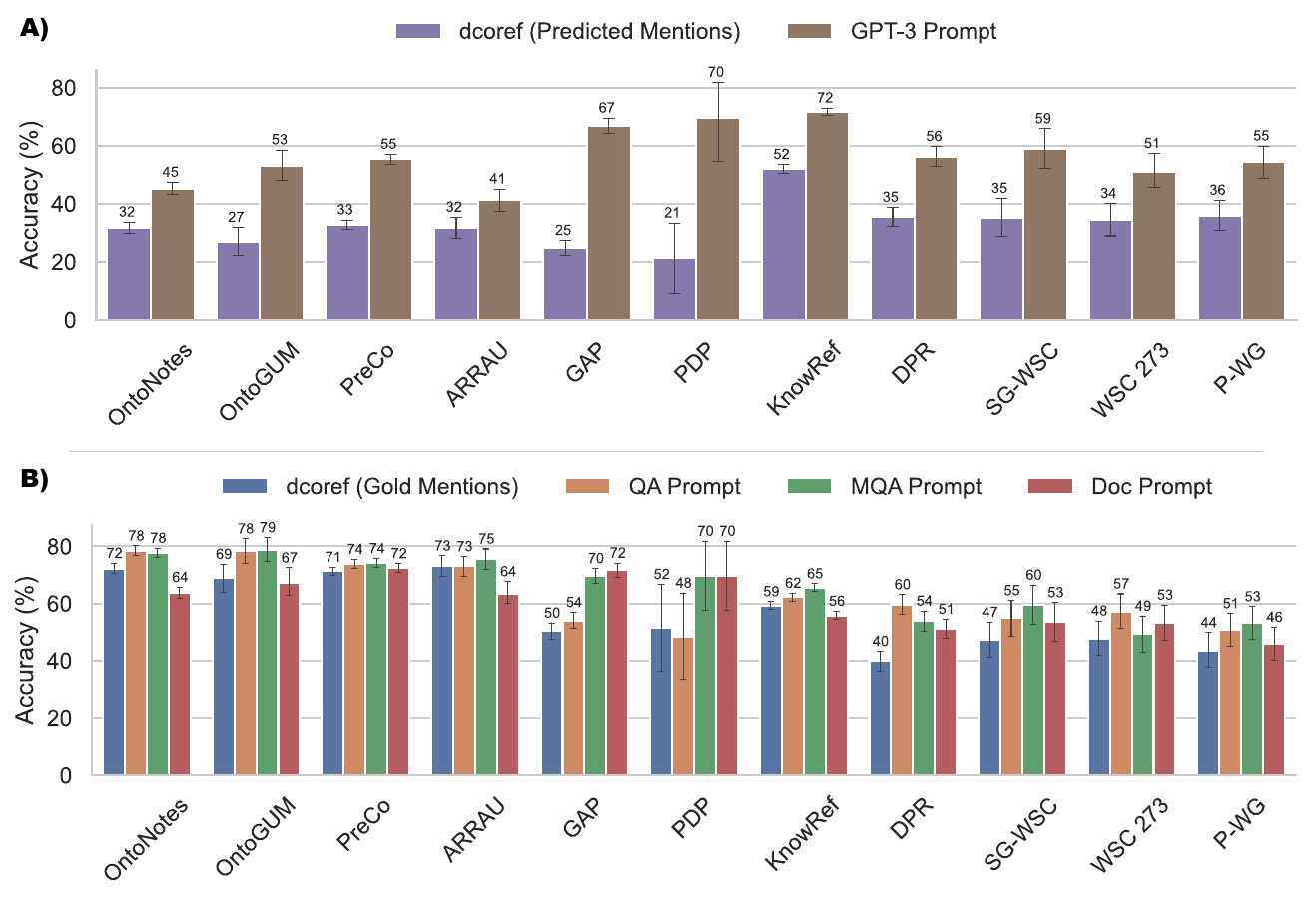}
  \caption{A comparison of the rule-based dcoref system~\citep{lee-etal-2013-deterministic} and the Llama 3.1 8B base model prompted for PCR using various prompts. \textbf{A)} Systems that do not need gold mention spans. Across datasets, Llama 3.1 with the GPT-3 prompt always outperforms the dcoref baseline. \textbf{B)} Systems that require gold mention spans as input. In general, prompted Llama 3.1 is more accurate than dcoref on both attested and constructed instances.
  }
  \label{fig:unsupervised-result}
\end{figure*}

\subsection{Datasets}
\label{sec:datasets}

We evaluate systems on 11 datasets including curated datasets of pronouns attested in natural corpora, such as OntoNotes \citep{ontonotes5}, and challenge sets of WSC-like instances, such as the original WSC test set~\citep{levesque2012winograd} and DPR~\citep{rahman-ng-2012-resolving}.

\subsubsection{Attested Pronominal Expressions}

As noted in the methods section (\S\ref{sec:method}), our tests that involve attested pronouns are based on restricting the set of annotations in more general coreference resolution datasets to be evaluated as binary classification problems similar to the WSC. The datasets that we use to achieve this are described below. Four of these are datasets of nominal identity coreference annotated in English-language, document-level passages (OntoNotes, OntoGUM, PreCo, and ARRAU). The remaining two focus exclusively on PCR (GAP and PDP).

\paragraph{OntoNotes} OntoNotes 5.0~\citep{ontonotes5} consists of seven genres including news, conversations, and web data annotated for coreference by two experts. This dataset has been used in prior work to explicitly evaluate PCR both in isolation~\citep{zhang-etal-2019-knowledge,zhang-etal-2021-brief,zhang-etal-2019-incorporating} as well as PCR as a failure case of more general coreference resolution systems~\citep{lu-ng-2020-conundrums}. We use the standard English CoNLL-2012 Shared Task version of this dataset~\citep{pradhan-etal-2012-conll}.

\paragraph{OntoGUM} OntoGUM~\citep{zhu-etal-2021-ontogum} is a reformatted version of the GUM corpus~\citep{gum-corpus} which was annotated for coreference by linguistic students. We use version 9.2.0 of OntoGUM. This dataset is designed to follow the same annotation guidelines as OntoNotes while expanding coverage to additional textual genres such as web forums and video blogs.

\paragraph{PreCo} PreCo \citep{chen-etal-2018-preco} is a large-scale dataset of English exams annotated for coreference.

\paragraph{ARRAU} ARRAU 2.1~\citep{arrau-corpus} is a dataset of written news and spoken conversations annotated for various anaphoric phenomena by experts.
We use the version formatted by \citet{xia-van-durme-2021-moving}. The annotation guidelines differ from OntoNotes, and additional phenomenon have been annotated including extensive semantic and syntactic features of mentions.

\paragraph{GAP} GAP~\citep{webster-etal-2018-mind} is a dataset of pronouns attested in English Wikipedia and annotated for coreference.
We study only instances where exactly one of two candidate antecedents is coreferring with the given pronoun to match our PCR problem formulation.

\paragraph{PDP} PDP~\citep{morgenstern2016planning} is a collection of 80 pronoun disambiguation problems attested in text which was used for the original version of the WSC in order to test systems on examples believed to be relatively easy.

\subsubsection{WSC-like Challenge Sets}

The five challenge sets that we evaluate on are as follows. To standardize the format of these datasets, we consider the lexical units $w_i$ to be syntactic words. We split the raw text into these syntactic words using the Stanza library~\citep{qi-etal-2020-stanza}.

\paragraph{KnowRef-60K} \citet{emami-etal-2020-analysis} presented WSC-like instances which were created by perturbing internet forum text using heuristic rules. Thus, these instances fall somewhere in between attested and constructed.

\paragraph{DPR} Definite pronoun resolution~\citep[DPR;][]{rahman-ng-2012-resolving} is a dataset of instances in a similar format to the original WSC without the strict requirement that instances cannot be resolved based on simple selectional preferences.

\paragraph{SuperGLUE WSC (SG-WSC)} The set of WSC instances used for the SuperGLUE benchmark~\citep{wang2019superglue} which was originally modified from WSC 273 and PDP.

\paragraph{WSC 273} The original WSC~\citep{levesque2012winograd} consisting of 273 instances. We manually annotated mentions to fit our format similar to as in \citet{McCann2018decaNLP} and \citet{toshniwal-etal-2021-generalization}.

\paragraph{Pronominal Winogrande (P-WG)} We use the portion of the Winogrande test set~\citep{sakaguchi2021winogrande} which contains person entities. We replace the underscore with an appropriate third-person pronoun as in \citet{porada2023investigating}.

\section{Results}
\label{sec:results}

We first present results comparing zero-shot prompting methods with the unsupervised dcoref system. We then compare the best performing prompting method against the supervised Maverick coreference resolution system and a supervised Llama 3.1 baseline. Across all figures, error bars represent 90\% confidence intervals. Results are presented on the corresponding test splits using the best model configuration. Additional details are presented in App.~\ref{sec:input-format}.

\paragraph{Comparing prompted LMs against earlier unsupervised systems, the challenge set assumption does hold.} Results for the fully unsupervised systems are presented in Figure~\ref{fig:unsupervised-result}. We observe that generally prompted LMs, which outperform dcoref on the WSC variants, also outperform dcoref on datasets of attested pronominal expressions. We also see that model performance is sensitive to the prompt format.

Exceptions are on the PDP dataset, whose small size makes it difficult to draw generalizable conclusions, and in the case of the Doc Prompt, which has high variance across datasets. \citet{le2023large} similarly found that Llama models did not consistently generalize with the Doc Prompt template.

\begin{figure}[ht]
\centering
  \includegraphics[width=0.45\textwidth]{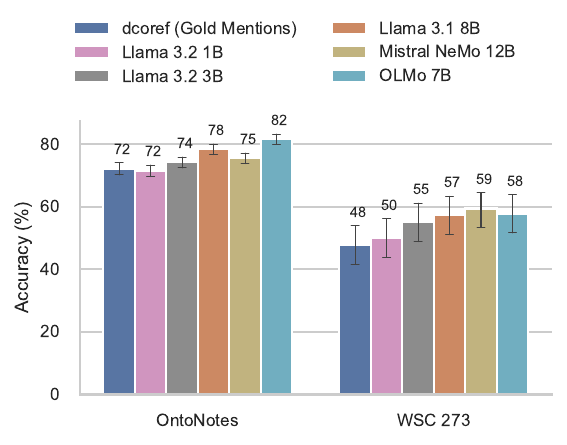}
  \caption{
  Accuracies of various LMs using the QA prompt template as compared against a dcoref baseline. We find that LMs generally outperform dcoref on both attested and constructed instances.
  }
  \label{fig:all-llms}
\end{figure}

In Figure~\ref{fig:all-llms}, we compare accuracies of various LMs using the QA prompt template. Our conclusion, that prompted LMs outperform dcoref on both constructed and attested instances, is consistent across LM families.

\begin{figure*}[t] 
\centering
  \includegraphics[width=0.99\textwidth]{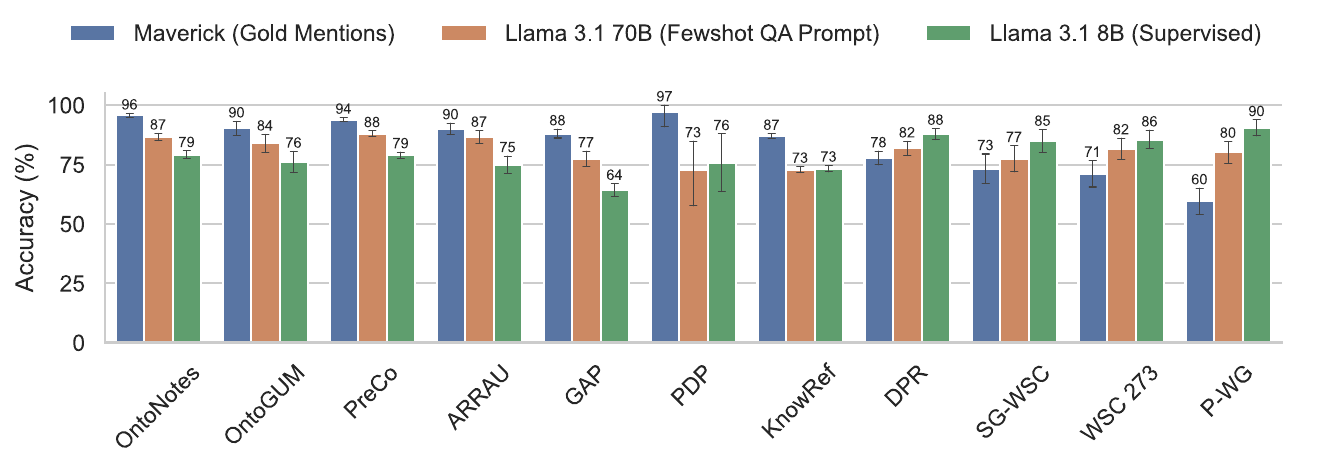}
  \caption{
  A comparison of the accuracy of Llama 3.1 70B instruct (32-shot) against the supervised Maverick coreference resolution system. We observe that the challenge set assumption does not hold; that is, despite being generally more accurate on WSC-like datasets, the prompted LM is less accurate on datasets of pronominal expressions attested in natural corpora. From left to right, the first six datasets consist of attested examples, and the remaining five are WSC-like challenge sets.
  }
  \label{fig:main-result}
\end{figure*}

\paragraph{However, when comparing prompted LMs against a supervised coreference resolution system, the challenge set assumption does not hold.}

In Figure~\ref{fig:main-result}, we present the accuracy of Llama 3.1 70B using the QA prompt in a few-shot setting compared against a supervised coreference resolution system. While the prompted LM is more accurate across  WSC-like datasets (with the exception of KnowRef), the supervised coreference resolution system is more accurate at resolving attested pronominal coreferences.

For this experiment, we consider the instruction-tuned version of Llama 3.1 with 32 in-context examples a prototypical example of an LM as evaluated on WSC-like datasets. When we compare against a supervised Llama 3.1 base model, trained on the Winogrande and DPR training sets for 5k steps, the difference in accuracies across datasets is even more extreme.

The exceptional case of KnowRef may be due to the fact that this dataset is constructed by perturbing attested pronominal expressions and may be overall more similar to collections of attested rather than constructed linguistic expressions.

\section{Analysis}
\label{sec:analysis}

Our results thusfar do not answer the question of \textit{why} the challenge set assumption does not hold. Heuristic estimates of features such as number and animacy are typically required to agree between an antecedent and a pronoun in order for the two to be coreferring, but these features are never required to resolve WSC instances per their design. Therefore, it may be the case that prompted LMs are not sufficiently considering these features for the attested PCR problems. To test this hypothesis, we analyze to what extent LMs could benefit from, or already incorporate implicitly, the use of such features. To do so, we experiment with oracle baselines including these features in the model input as a verbalized statement.

\subsection{Verbalized Features}
\label{sec:verbalized-features}

The verbalized features that we consider are those annotated in the ARRAU corpus. These are: 1) grammatical gender, 2) number, 3) enamex type (i.e., semantic type: is the entity a person, organization, or location?), and 3) distance between mentions. We also explore incorporating a gold, annotated label as an oracle baseline.

\paragraph{Prompt.}

Verbalized features are appended to the input string in the form:

\begin{quote}
    \small
    The [FEATURE\_NAME] of ``[X]'' is [Y].
\end{quote}

For example, the passage in Figure~\ref{fig:formulation-example} is prepended with verbalizations such as:

\begin{quote}
    \small
    The grammatical gender of ``Jim'' is male.
\end{quote}

Results of this experiment are presented in Table~\ref{tab:added-features-results}. We observe some accuracy increase from the inclusion of grammatical gender, but otherwise no influence. Meanwhile, the oracle baseline suggests models are capable of incorporating verbalized features that perfectly align with the correct antecedent prediction (i.e., gold labels).

\begin{table}[th]
\small
\centering
\begin{tabular}{lr}
\toprule
                                           & ARRAU \\
                                           \midrule
Llama 3.1 70B (QA Prompt) & 0.86                       \\
+ gold gender                                                & 0.87                       \\
+ gold number                                                & 0.85                       \\
+ gold enamex type                                           & 0.86                       \\
+ distance between mentions                                  & 0.84                       \\
+ gold label (oracle)                                        & 0.99                       \\
\bottomrule
\end{tabular}
\caption{Results including additional features in the model input on the ARRAU validation set.}
\label{tab:added-features-results}
\end{table}

\section{Ensembling Systems for Better Performance}
\label{sec:ensembling}

Finally, we present results for our proposed ensembling method. This method is motivated as follows: because the challenge set assumption does not hold, prompted LMs and task-specific systems have distinct strengths at PCR.

\subsection{Method}

This method functions by heuristically determining if $x$ corresponds to a salient discourse entity, in which case a supervised coreference resolution system is used to predict the correct antecedent $a$. Otherwise, $x$ is resolved using a prompted LM. This approach benefits from the fact that supervised coreference resolution systems are relatively accurate at resolving pronominal expressions that corefer to the most salient discourse entities. Meanwhile, prompted LMs are relatively accurate at resolving pronominal expressions referring to infrequently mentioned entities.

\subsection{Implementation}

For our proposed ensembling method, we first predict pronominal coreferences using both the supervised Maverick system and the prompted Llama 3.1 70B instruct model as before. We then heuristically determine if a candidate antecedent corresponds to a salient discourse entity based on the number of coreferring noun phrases predicted by Maverick in an end-to-end setup. When the number of predicted coreferring mentions is greater than two (that is, the pronoun is estimated to corefer to more than one other linguistic expression) we use the Maverick predictions given gold mention spans. Otherwise, we use the Llama predictions.

\subsection{Results}

\begin{table}[t]
\small
\centering
\begin{tabular}{lccc}
\toprule
System & OntoGUM & PreCo & WSC 273  \\
\midrule
Maverick & 0.90	& 0.94 & 0.71 \\
Llama 3.1 70B & 0.84 & 0.88 & 0.82 \\
\midrule
Ensemble & 0.90	& 0.95 & 0.84\\
\bottomrule
\end{tabular}

\caption{
  Accuracy of the ensemble method compared against Maverick (supervised coreference resolution) and prompted Llama 3.1 70B instruct.
  }
  \label{tab:ensemble-results}
\end{table}

We present the results for our ensembling method on three out-of-domain datasets of attested pronominal expressions in Table~\ref{tab:ensemble-results}. The ensemble predictions are at least as accurate as the best performing model, and in the case of PreCo and WSC 273, more accurate than the single most accurate system. Results across all datasets are presented in App.~\ref{sec:full-ensemble}.

\section{Discussion}
\label{sec:discussion}

Coreference resolution systems have traditionally struggled at resolving pronouns when the resolution depends on semantic knowledge related to high-order predicate-argument relations~\citep{kehler-etal-2004-non,durrett-klein-2013-easy,zhang-etal-2019-sp}. Meanwhile, our results suggest that resolving WSC instances, which are designed to explicitly rely on such knowledge, is in some ways relatively easier than other cases for prompted LMs. Therefore, our intuitions as a research community regarding what constitutes challenging examples may not always be aligned with the actual failure cases of newer modeling paradigms. Consequently, we must be careful as a community to not interpret high performance on challenge sets as indicating that the more general task being studied can consistently be solved by a given system.

\paragraph{The Solvability of PCR.}

Our experiments and results are not intended to make claims regarding the solvability of the task of PCR. It may be that alternative prompting formats exist for which Llama models are relatively more accurate at resolving attested pronominal coreferences, and one would expect accuracy to increase with LM size. What our results do show, rather, is that existing approaches that are successful on the WSC and variants cannot generalize to all attested PCR problems.

\paragraph{Coreference and Substitutability.}

By their design, WSC instances can be formatted as a cloze-style task where the correct antecedent is that which is most likely to be substituted for the ambiguous pronoun. Substitutability and coreference are related but distinct concepts, however. While WSC instances are difficult in that they cannot be solved with the agreement of features between a pronoun and a candidate antecedent, they differ from some attested PCR problems in that for WSC instances the concept of coreference is aligned with substitutability. One possible hypothesis to explain our results is that this alignment is useful for solving the WSC. This hypothesis is based on the idea that substitutability can be formatted as a cloze-style task and is therefore closely aligned with the LM pretraining objective.

\paragraph{Data Contamination.}

An open question is whether LMs are exposed to the WSC or other datasets' test instances during pretraining. \citet{elazar2023s} estimate that up to 30\% of WSC test instances may be contaminated in the training corpus of Llama and other language models. However, OntoNotes, OntoGUM, Winogrande, Knowref, and GAP are estimated to have close to zero contamination according to the Data Contamination Database~\citep{data-contamination-database}. ARRAU is not publicly distributed and also unlikely to be contaminated. Because our results are consistent for datasets that are likely not contaminated, we believe that issues of data contamination are unlikely to invalidate our findings.

\section{Conclusion}
\label{sec:conclusion}

The ability to disambiguate pronominal expressions is necessary for interpreting natural language and has been used extensively as a benchmark to evaluate models of semantics and discourse.

In this work, we study several possible approaches to modeling pronominal coreference. Across evaluations, we find that prompting a large language model (LM) outperforms other approaches on the WSC, but underperforms on certain attested occurrences of pronouns annotated for coreference in OntoNotes and related datasets.

\section{Limitations}
\label{sec:limitations}

We focus on a limited formulation of PCR. One could expand on the scope of these results by considering additional formulations of PCR as well as additional types of pronominal or other proform expressions (\textit{e.g.,} a broader set of expressions considered as pronouns such as first and second person or reflexive pronouns). Additionally, the scope of coreference could be more explicitly specified by distinguishing identity coreference versus other related phenomenon such as binding.

We also did not consider differences between dataset annotation in detail. For example, datasets differ in the annotation of mention spans. Our experiments using gold mention annotations provide some insight into the impact of these differences, but this impact could be studied more thoroughly by explicitly considering how mention spans are annotated within each dataset.

Furthermore, we did not consider model failure cases in detail beyond our ablation experiments on the ARRAU dataset. For example, how performance might vary based on genre and how this differs between systems. For instance, LinkAppend was trained with genre and speaker metadata.

Finally, expanding our evaluations to multilingual pronominal anaphora and subsets of coreference datasets other than the English language would allow for new results regarding phenomenon that are more prominent outside of English (\textit{e.g.,} zero-anaphors) or do not exist in English (\textit{e.g.,} switch reference and obviation).

\section{Ethics Statement}
\label{sec:ethics-statement}

PCR systems are known to perform disparately on subgroups which has ethical implications particularly for potential real-world use cases~\citep{zhao-etal-2018-gender,rudinger-etal-2018-gender,webster-etal-2018-mind,hossain-etal-2023-misgendered}. We therefore do not recommend or endorse the use of these systems for downstream purposes such as real-world, commercial applications; rather, our experiments are focused solely on the validity of certain assumptions of existing challenge sets.

\section*{Acknowledgements}

Ian Porada is supported by a doctoral fellowship from the Fonds de Recherche du Québec Nature et Technologies (FRQ-NT). This research was enabled in part by compute resources provided by Mila (mila.quebec). We thank the anonymous reviewers for their valuable suggestions which greatly improved this paper.

\bibliography{anthology,custom}

\appendix

\section{Differences in formulations of PCR}
\label{sec:evaluation-differences}

Our formulation of PCR follows the precise setup proposed by \citet{zhang-etal-2021-brief} which was in turn based on earlier formulations which also considered fixed subsets of English pronouns in restricted contexts. These restrictions were viewed as reasonable because most antecedents occur within the local context of a pronoun; e.g., \citet{yang-etal-2003-coreference} observed that the antecedent is within the local context 95\% of the time in the MUC corpus.

We similarly formatted WSC-like challenge sets in this way to allow for a fair comparison. For instance, WSC-like datasets may initially contain pronominal expressions outside our considered set such as \textit{one} and \textit{y'all}.

For additional details, we release our preprocessing code at \url{https://github.com/ianporada/challenge-set-assumption}.

\subsection{Additional Considerations}

In this section, we outline differences in how existing work has approached PCR and which choices we make in setting the scope of our analysis.

There is a tradeoff between evaluating all forms of pronominal coreference that might occur in natural language and evaluating those forms that have been identified and defined in such a way that they can be reliably annotated in existing corpora. With this perspective, our goal is more oriented towards the latter. That is, we do not intend to analyze all conceivable coreferences of all possible pronominal expressions. Rather, we take the intersection of existing work to better understand how well models generalize across datasets.

\paragraph{End-to-end v.s. mention-linking:} As an end-to-end task, the goal of PCR is to determine with which linguistic expressions a pronoun corefers given only the raw context and identification of the pronoun. In contrast, it could be the case that candidate antecedent mentions are already identified, in which case the task of PCR consists of resolving the correct antecedent. Common approaches are to score each candidate independently or pairwise~\citep{yang-etal-2008-twin}. We compare existing systems within the category with which they can perform the task.

\paragraph{One v.s. many mentions:} A discourse entity can be realized as multiple coreferring linguistic expressions in a discourse. These realizations form a coreference cluster. In the case that multiple realizations appear in the context of a pronominal expression, there are multiple possible interpretations for what is considered the correct antecedent to be resolved to the pronoun. Popular approaches are to consider the most recent mention \citep{liang-wu-2003-automatic} or any one of the coreferring mentions as a valid antecedent. We consider the most recent mention to be the valid antecedent which is the approach most commonly taken in existing work~\citep{zhang-etal-2021-brief}. Nonetheless, we do not consider instances where multiple coreferring expressions appear within the immediate context $w$ to allow for our binary classification evaluation.

\paragraph{Mention boundaries:} Finally, datasets differ in the annotation of mention boundaries~\citep{moosavi-etal-2019-using}. For example, the antecedent noun phrase ``a young man'' in Figure~\ref{fig:intro-example} is annotated in the dataset as ``man'' whereas in OntoNotes would be annotated as the maximal dominating span ``a young man of large fortune from the north of England'' according to the annotation guidelines. In the case where a PCR model functions end-to-end a reasonable assumption might be that the pronoun should corefer with at least a mention containing the head word of the correct antecedent and no mention containing the head word of the incorrect antecedent~\citep{crac-2022-crac}; however, optimizing for head words in this way has been shown to lead to strange modeling design decisions that do not align with human intuition~\citep{crac-2023-crac}. \citet{moosavi-etal-2019-using} presents a method for normalizing mention boundaries so some minimal span which is a reasonable choice for end-to-end systems and may be useful for future work. For mention-linking, we simply consider the dataset's annotated mention. We do not consider more complex phenomenon such as split-antecedents and discontinuous mentions in our analysis, but these would also be interesting to investigate in future work.

\begin{figure*}[t] 
\centering
  \includegraphics[width=0.99\textwidth]{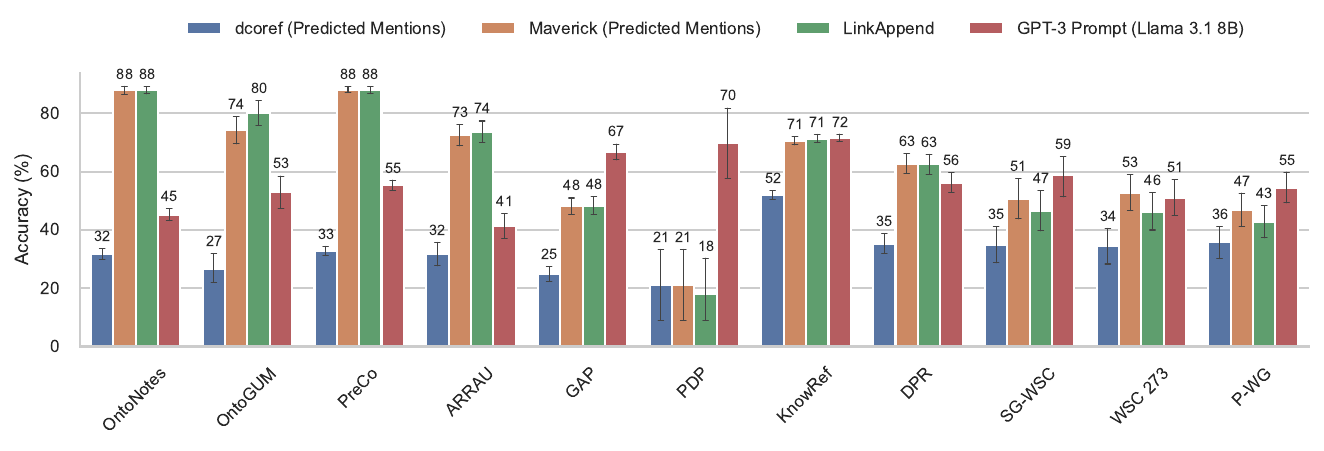}
  \caption{
  A comparison of the LinkAppend coreference resolution system and other systems that also do not rely on gold mention annotations. The GPT-3 Prompt is most accurate on certain WSC datasets, but performance is near random chance on OntoNotes, OntoGUM, PreCo, and ARRAU. This may be due to the difficulty in predicting the appropriate spans of linguistic expressions.
  }
  \label{fig:linkappend}
\end{figure*}

\section{Input Format}
\label{sec:input-format}

In this section we provide more details regarding how the input was formatted. We also discuss additional results given other formats. We find that our conclusions are consistent across these decisions.

\paragraph{Speaker Information.} Many datasets of coreference annotations consist of spoken language and include corresponding speaker metadata. We tested models both with and without this metadata and report models in their best configuration. We found that Maverick and dcoref perform best with speaker information, which including speaker data in LLM input in the form of ``SPEAKER\_NAME: ...'' had marginal negative effect on LMs' performance.

\paragraph{Input Length.} The datasets of curated natural corpora that we consider typically consist of relatively long document contexts. Therefore, we experimented with including in the input only the local context $w$ or the entire document. We find that including the full context length has a marginal effect on performance. In the unsupervised case, we use only the local context $w$ whereas for supervised models we include the full document in the input for all models (both finetuned systems and few-shot LMs).

\paragraph{dcoref} For the dcoref baseline, we do not use gold parses (since not all datasets include parse information) and rather use parses predicted by the Stanford CoreNLP pipeline.

\paragraph{LinkAppend} As an additional representative example of supervised systems, we consider the state-of-the-art LinkAppend coreference resolution system \citep{bohnet-etal-2023-coreference}. We use the publicly released weights of the best performing system which consists of the multilingual mT5-XXL language model (13B params) finetuned on OntoNotes. Results for the LinkAppend system without gold mention spans are presented in Figure~\ref{fig:linkappend}.

\begin{table*}[t]
\small
\centering
\begin{tabular}{lcccccccccc}
\toprule
System & ON & OG & PreCo & ARRAU & GAP & KnowRef & DPR & SG-WSC & WSC 273 & P-WG \\
\midrule
Maverick & \textbf{0.96}\makebox[0pt][l]{$^*$} & \textbf{0.90}	& 0.94 & 0.90 & 0.88 & \textbf{0.87}\makebox[0pt][l]{$^*$} & 0.78 & 0.73 & 0.71 & 0.60 \\
Llama 3.1 70B & 0.87 & 0.84 & 0.88 & 0.87 & 0.77 & 0.73 & 0.82 & 0.77 & 0.82 & 0.80 \\
\midrule
Ensemble & 0.93 & \textbf{0.90}	& \textbf{0.95} & \textbf{0.91} & \textbf{0.91}\makebox[0pt][l]{$^*$} & 0.77 & \textbf{0.85} & \textbf{0.79} & \textbf{0.84} & \textbf{0.81} \\
\bottomrule
\end{tabular}

\caption{
  Accuracy of the ensemble method compared against Maverick (supervised coreference resolution) and prompted Llama 3.1 70B instruct. ON denotes OntoNotes and OG denotes OntoGUM. The most accurate model in each column is marked in bold. * indicates results that are statistically significant as compared to the next best model based on a t-test with p-value $< 0.1$.
  }
  \label{tab:full-ensemble-results}
\end{table*}

\section{Full Ensemble}
\label{sec:full-ensemble}

Results of the ensembling method across all datasets are presented in Table~\ref{tab:full-ensemble-results}. (We do not consider PDP due to its small test set size.) The ensembled predictions outperform any single system on all datasets except OntoNotes and KnowRef. The OntoNotes test set is known to have a high lexical overlap with the training set which could possibly explain the exceptional superior performance of the supervised Maverick model on this dataset. To test if this is the case, we can also consider the public Maverick weights trained on PreCo in the same setup in which case the Maverick model accuracy is 0.71 and is significantly outperformed by the ensemble approach (0.89 in this case).

In the case of KnowRef, it is not clear why the ensemble approach is less accurate than the supervised system in contrast to all other datasets. This may be related to the relatively poor accuracy of Llama 3.1 on KnowRef and would be interesting to investigate in future work.

\section{Dataset Details}
\label{sec:dataset-details}

\subsection{Examples}

Here we present example instances from the validation sets of those datasets that include a validation split. For readability, we show only the local context $w$.

\begin{figure}[H]
\centering
  \includegraphics[width=0.475\textwidth]{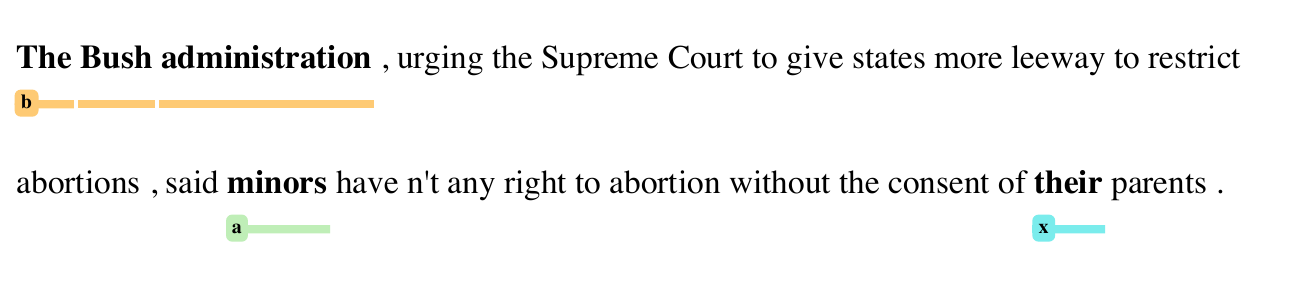}
  \caption{
  An instance from the OntoNotes dataset.
  }
  \label{fig:ontonotes-example}
\end{figure}

\begin{figure}[H]
\centering
  \includegraphics[width=0.475\textwidth]{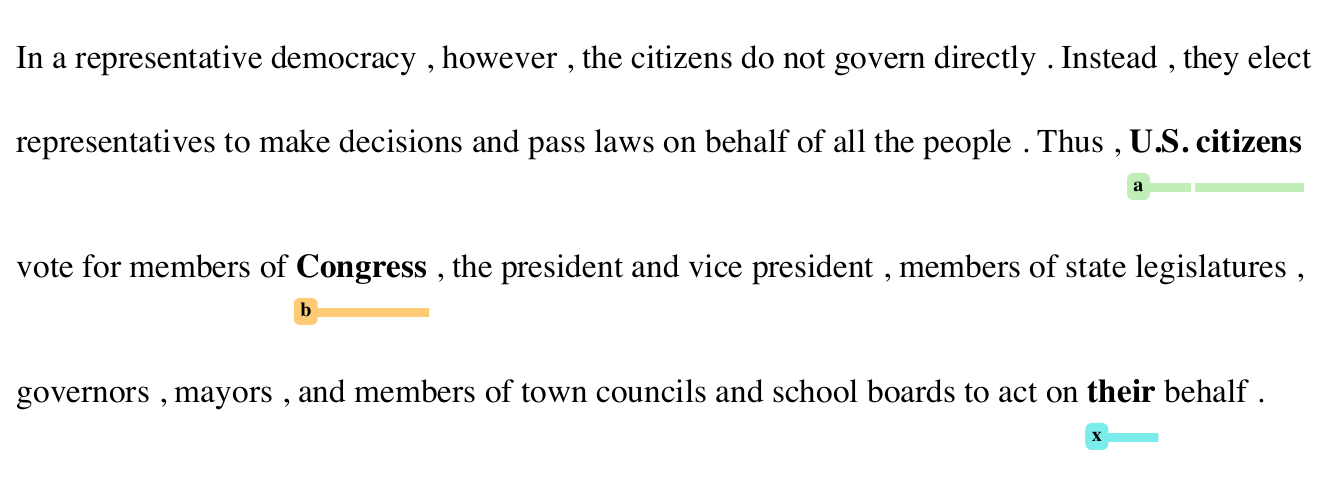}
  \caption{
  An instance from the OntoGUM dataset.
  }
  \label{fig:ontogum-example}
\end{figure}

\begin{figure}[H]
\centering
  \includegraphics[width=0.475\textwidth]{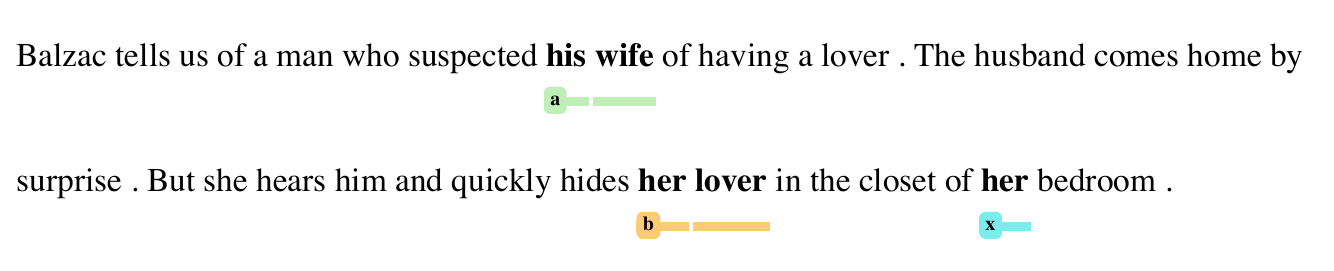}
  \caption{
  An instance from the PreCo dataset.
  }
  \label{fig:preco-example}
\end{figure}

\begin{figure}[H]
\centering
  \includegraphics[width=0.475\textwidth]{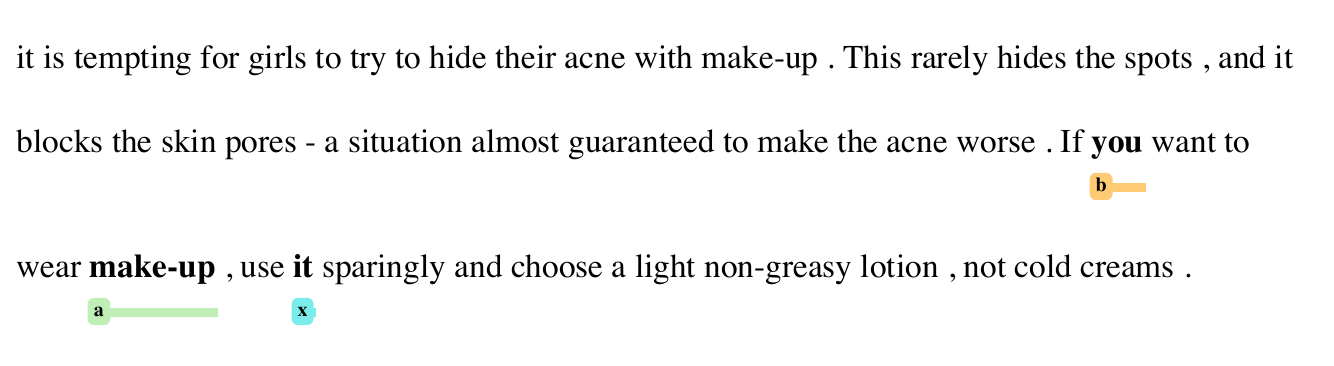}
  \caption{
  An instance from the ARRAU dataset.
  }
  \label{fig:arrau-example}
\end{figure}

\begin{figure}[H]
\centering
  \includegraphics[width=0.475\textwidth]{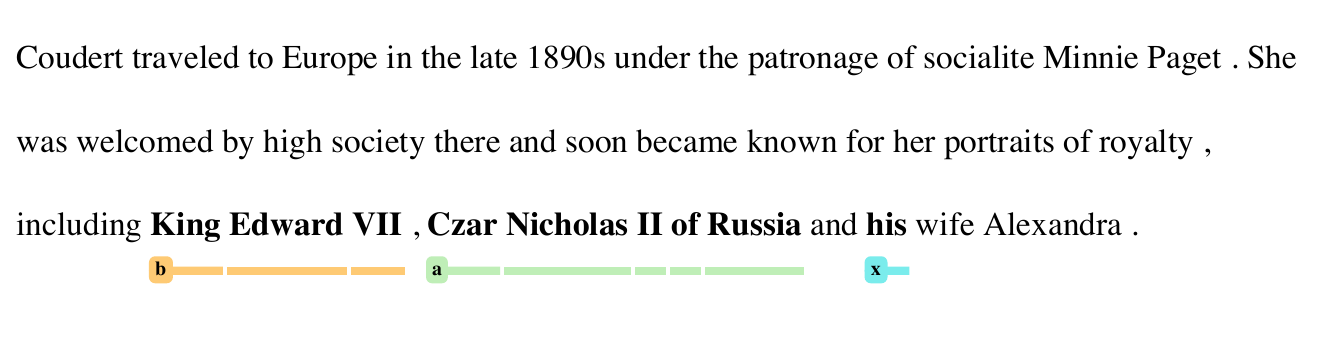}
  \caption{
  An instance from the GAP dataset.
  }
  \label{fig:gap-example}
\end{figure}

\begin{figure}[H]
\centering
  \includegraphics[width=0.475\textwidth]{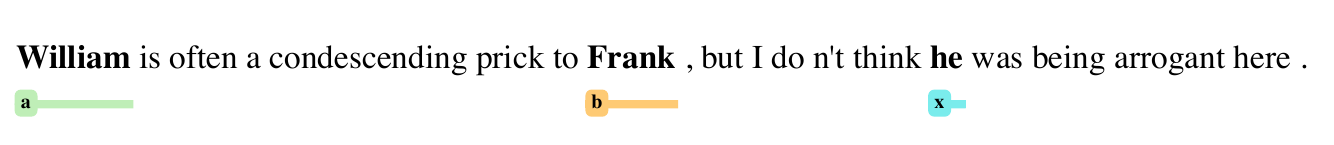}
  \caption{
  An instance from the KnowRef-60K dataset.
  }
  \label{fig:knowref-example}
\end{figure}

\begin{figure}[H]
\centering
  \includegraphics[width=0.475\textwidth]{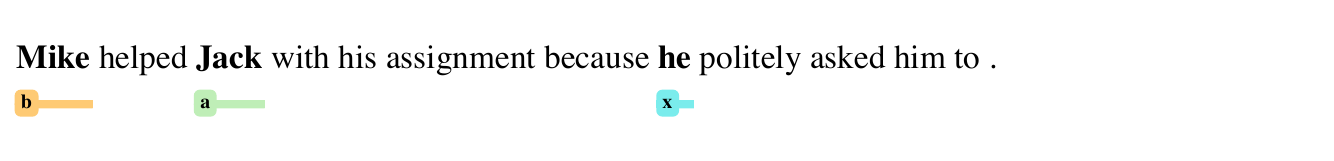}
  \caption{
  An instance from the DPR dataset.
  }
  \label{fig:dpr-example}
\end{figure}

\begin{figure}[H]
\centering
  \includegraphics[width=0.475\textwidth]{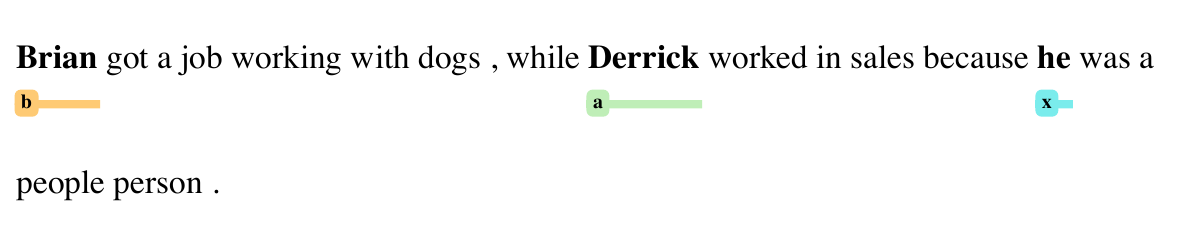}
  \caption{
  An instance from the Pronominal Winogrande (P-WG) dataset.
  }
  \label{fig:winogrande-example}
\end{figure}

\subsection{Summary Statistics}

\paragraph{OntoNotes}

\begin{itemize}
    \item License: LDC User Agreement for Non-Members
    \item Final validation instances: 1,536
    \item Final test instances: 1,642
\end{itemize}

\paragraph{OntoGUM}

\begin{itemize}
    \item License: Varies by subcorpus. All annotations are \texttt{cc-by-4.0}
    \item Final validation instances: 272
    \item Final test instances: 236
\end{itemize}

\paragraph{PreCo}

\begin{itemize}
    \item License: None specified
    \item Final validation instances: 2,167
    \item Final test instances: 2,248
\end{itemize}

\paragraph{ARRAU}

\begin{itemize}
    \item License: LDC User Agreement for Non-Members
    \item Final validation instances: 179
    \item Final test instances: 411
\end{itemize}

\paragraph{GAP}

\begin{itemize}
    \item License: \texttt{apache-2.0}
    \item Final validation instances: 203
    \item Final test instances: 832
\end{itemize}

\paragraph{PDP}

\begin{itemize}
    \item License: \texttt{cc-by-4.0}
    \item Final test instances: 33
\end{itemize}

\paragraph{KnowRef-60K}

\begin{itemize}
    \item License: \texttt{cc-by-4.0}
    \item Final validation instances: 21,240
    \item Final test instances: 3,061
\end{itemize}

\paragraph{DPR}

\begin{itemize}
    \item License: None specified
    \item Final test instances: 558
\end{itemize}

\paragraph{SuperGLUE WSC}

\begin{itemize}
    \item License: Custom (research usages)
    \item Final test instances: 146
\end{itemize}

\paragraph{WSC 273}

\begin{itemize}
    \item License: \texttt{cc-by-4.0}
    \item Final test instances: 180
\end{itemize}

\paragraph{Pronominal Winogrande}

\begin{itemize}
    \item License: \texttt{cc-by-4.0}
    \item Final test instances: 209
\end{itemize}

\end{document}